\documentclass[runningheads]{llncs}
\usepackage{graphicx}
\usepackage{bbding}
\usepackage[table,xcdraw]{xcolor}
\usepackage{hyperref}
\usepackage{multirow}
\usepackage{booktabs}
\usepackage{multirow}
\usepackage[table]{xcolor}
\usepackage{colortbl}
\usepackage{array}
\usepackage{wrapfig}

\begin{document}
\title{MedTri: A Platform for Structured Medical Report Normalization to Enhance Vision–Language Pretraining}

\author{Yuetan Chu\inst{1} \and
Xinghua Ma\inst{2} \and
Xinran Jin\inst{2} \and
Gongning Luo\inst{2(}\Envelope\inst{)} \and
Xin Gao\inst{1(}\Envelope\inst{)}}

\institute{King Abdullah University of Science and Technology (KAUST), Thuwal, Saudi Arabia\\
\email{xin.gao@kaust.edu.sa} \and
Faculty of Computing, Harbin, China.\\
\email{luogongning@hit.edu.cn}}

\maketitle              
\begin{abstract}
Medical vision–language pretraining increasingly relies on medical reports as large-scale supervisory signals; however, raw reports often exhibit substantial stylistic heterogeneity, variable length, and a considerable amount of image-irrelevant content. Although text normalization is frequently adopted as a preprocessing step in prior work, its design principles and empirical impact on vision–language pretraining remain insufficiently and systematically examined. In this study, we present \textbf{MedTri}, a deployable normalization framework for medical vision–language pretraining that converts free-text reports into a unified \textit{[Anatomical Entity: Radiologic Description + Diagnosis Category]} triplet. This structured, anatomy-grounded normalization preserves essential morphological and spatial information while removing stylistic noise and image-irrelevant content, providing consistent and image-grounded textual supervision at scale. Across multiple datasets spanning both X-ray and computed tomography (CT) modalities, we demonstrate that \emph{structured, anatomy-grounded text normalization is an important factor in medical vision–language pretraining quality}, yielding consistent improvements over raw reports and existing normalization baselines. In addition, we illustrate how this normalization can easily support targeted text-level augmentation strategies, including knowledge enrichment and anatomy-grounded counterfactual supervision, which provide complementary gains in robustness and generalization without altering the core normalization process. Together, our results position structured text normalization as a critical and generalizable preprocessing component for medical vision–language learning, while MedTri provides this normalization platform. Code and data will be released at \texttt{https://github.com/Arturia-Pendragon-Iris/MedTri}.

\end{abstract}

\section{Introduction}
\vspace{-0.75em}
Vision--language pretraining (VLP) that leverages naturally paired medical images and medical reports has emerged as a powerful paradigm in medical image analysis~\cite{mimic_xray}\cite{clip}. Such pairs provide large-scale semantic supervision without additional annotation effort~\cite{vlm_pretrain_1}\cite{vlm_pretrain_2}\cite{vlm_pretrain_3}, as reports contain expert interpretations grounded in patient-specific visual findings. By jointly modeling visual content and textual descriptions, medical VLP strategies have demonstrated strong capabilities in capturing disease-relevant semantics, improving representation quality, and enhancing generalization across diverse downstream tasks~\cite{medfilip}. This synergy positions vision-language alignment as a foundational component for modern medical image pretraining.

\begin{figure}[!t]
\centerline{\includegraphics[width=\textwidth]{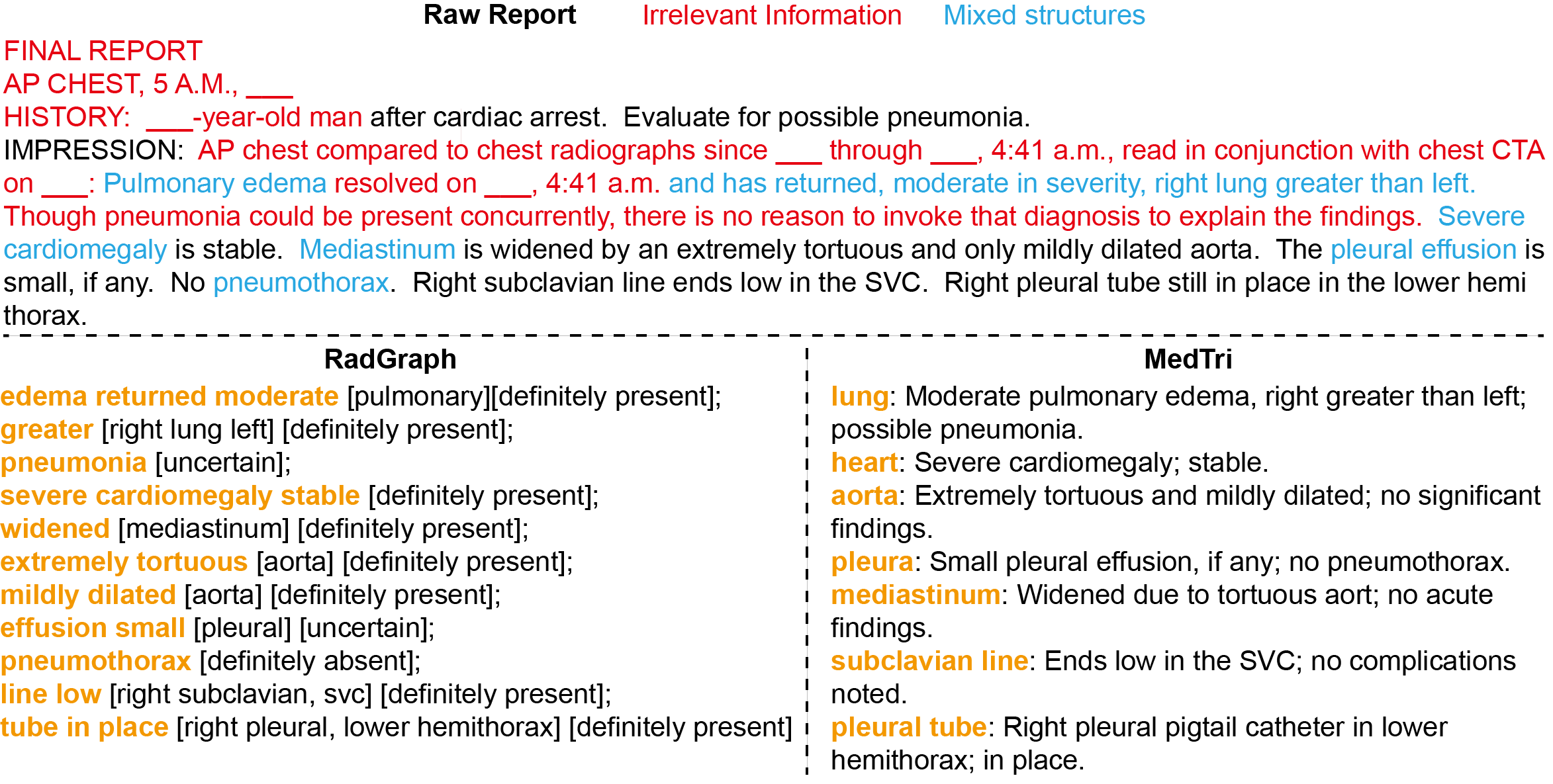}}
	\caption{Illustration of normalization differences across raw reports, RadGraph, and MedTri. The raw clinical report contains irrelevant content and mixed structures. RadGraph extracts only diagnostic entities with limited imaging descriptions. MedTri produces anatomically anchored, image-grounded triplets that preserve morphological and spatial detail while removing stylistic and clinically irrelevant text.}
 \vspace{-1.5em}
	\label{example}
\end{figure}

Despite the success of VLP, the textual supervision provided by raw clinical reports introduces several obstacles for effective multimodal alignment. Medical reports vary widely in style, verbosity, and structure, often mixing image-grounded observations with unrelated clinical history or management recommendations~\cite{parrot}. This heterogeneity reduces image-relevant signals, inflates sequence length, and weakens the fine-grained correspondence between visual findings and textual descriptions. As a result, recent studies increasingly adopt text normalization to standardize report content and enhance the stability of vision-language learning~\cite{medfilip}\cite{ct_glip}. However, text normalization is often introduced as a preprocessing component and combined with other architectural designs, while its standalone design principles and empirical impact on vision–language pretraining remain insufficiently examined. Moreover, existing normalization, such as schema-based or NER-driven systems (e.g., RadGraph~\cite{radgraph}), primarily focus on entity extraction, whereas cloud-based LLM methods rely on large-scale generative rewriting at the expense of increased computational overhead and potential privacy concerns. Consequently, the field still lacks a lightweight, anatomically expressive, and locally deployable normalization solution.

In this study, we present \textbf{MedTri}, a structured normalization platform for medical vision–language pretraining. MedTri converts free-text radiology reports into a unified \textit{[Anatomical Entity: Radiologic Description + Diagnosis Category]} triplet that preserves essential morphological and spatial information while removing stylistic noise and image-irrelevant content~\cite{Feng_2024_clipcleaner}~\cite{Feng_2024_NoiseBox}. This design enables consistent, efficient, and privacy-preserving report normalization suitable for large-scale pretraining. Beyond the normalization itself, MedTri further provides a modular interface that allows additional text-level augmentation on top of the normalized triplet. Using this interface, we instantiate two optional examples: knowledge enrichment and anatomy-grounded counterfactual augmentation. Across multiple datasets covering both X-ray and CT modalities, we systematically demonstrate that \emph{structured, anatomy-grounded text normalization is an important factor in medical vision–language pretraining quality}, with MedTri consistently outperforming raw reports and existing normalization approaches. The optional augmentation modules provide further, complementary gains in performance and generalization. Together, these results position MedTri as a practical and anatomically expressive normalization platform for medical vision–language learning.

\begin{table}
\vspace{-1em}
\centering
\resizebox{0.9\textwidth}{!}{
\normalsize
\begin{tabular}{l|c|c|c}
\hline
\textbf{Dataset} & \textbf{\# Reports} & \textbf{Image Modality} & \textbf{Body Region} \\
\hline
\rowcolor{gray!10}
MIMIC-CXR~\cite{mimic_xray}     & 65,000  & X-ray               & Chest \\
CT-RATE~\cite{CTRATE}       & 20,000  & CT                  & Chest \\
\rowcolor{gray!10}
CT-INSPECTION~\cite{inspect} & 10,000  & CT                  & Chest \\
AbdomenAtlas~\cite{abdomen}  & 5,000   & CT                  & Abdomen \\
\rowcolor{gray!10}
Parrot~\cite{parrot}        & 2,658   & CT, MRI, X-ray, etc & Head, abdomen, chest, etc \\
\hline
\end{tabular}} 
\caption{Overview of the report dataset used to develop and evaluate the MedTri normalization platform.}
\vspace{-2em}
\label{dataset}
\end{table}

\section{Method}
\vspace{-0.75em}
\subsection{Structured Triplet}
To support stable normalization for medical vision-language pretraining, we adopt a structured triplet that decomposes each report into a set of clinically grounded triplets:
$$[\textit{Anatomical Entity:}\ \textit{Radiologic Description} + \textit{Diagnosis Category}]$$

The schema captures the minimal semantic unit of radiologic reasoning, consisting of an anatomical anchor, objective imaging attributes, and an associated diagnostic interpretation when present. By explicitly disentangling these components, MedTri converts heterogeneous free-text reports into anatomy-level alignment units, reducing lexical and stylistic variability while preserving semantically discriminative, image-grounded information for vision-language pretraining.

\subsection{Local Model Development}
Clinical reports from different institutions exhibit substantial variability in style, grammar, and diagnostic phrasing. To ensure that MedTri remains robust to this heterogeneity, we established a dataset with more than 100,000 reports selected from multiple publicly available datasets covering diverse imaging modalities and anatomical regions (Table~\ref{dataset}). Using this dataset, we first generated structured reference summaries through a standardized ChatGPT-5.1 prompt specifically designed for our triplet schema. The prompt was iteratively refined on approximately 1,000 cases and then applied to the full dataset, yielding paired samples ${(x_i, y_i)}$ in which $x_i$ is the original report and $y_i$ is its normalized triplet (Table~\ref{prompt})~\cite{lu2023llm}.

To enable scalable and privacy-preserving deployment, we train a lightweight biomedical encoder–decoder model (BioBart~\cite{BioBART}) to approximate the cloud LLM-distilled supervision. The model is fine-tuned on the paired dataset using standard cross-entropy loss, following a traditional Seq2Seq formulation~\cite{text_pre_1}\cite{text_pre_2}. This locally deployable text-transfer model forms the core of the MedTri platform, allowing rapid and consistent normalization without relying on large cloud-based language models. A total of 500 reports were randomly sampled from the established dataset for model testing, whereas the remaining reports were used for model training and validation.

\begin{table}[h!]
\centering
\begin{tabular}{|p{1.5cm}|p{10.5cm}|}
\hline
\textbf{Input:} &
IMPRESSION: 1. No evidence of pulmonary embolism. No acute intrathoracic pathology. 
2. Mild bronchial thickening, nonspecific, but may reflect infectious or inflammatory process. 
3. Main pulmonary artery is dilated measuring 39 mm, possibly reflecting background pulmonary hypertension. \\ 
\hline

\textbf{Prompt:} &
You will receive one paragraph containing radiology findings (image-based description) and a final diagnosis. Your ONLY task is to convert this information into a structured radiology summary using the specified format.\newline
Each line must follow this exact format: [Anatomical Term]: [Radiologic Description + Diagnosis Category].\newline
List only the anatomical regions mentioned in the findings or diagnosis. Maintain the anatomical hierarchy (e.g., lung before airway, heart before aorta) if multiple structures are present.\newline
For each anatomical term, include radiologic findings (location, size, attenuation, density, texture, etc.) and diagnosis category if explicitly stated.\newline
Use objective imaging language only. Use direct descriptive phrasing. Keep each anatomical line concise and single-sentence, using semicolons to separate multiple findings.
\\
\hline

\textbf{Output:} &
Lung: No evidence of pulmonary embolism; no acute intrathoracic pathology.\\
&Bronchi: Mild bronchial thickening; nonspecific; may reflect infectious or inflammatory process.\\
&Main pulmonary artery: Dilated; measures 39 mm; possibly reflecting background pulmonary hypertension.\\
\hline
\end{tabular}

\caption{Example of the standardized ChatGPT-5.1 prompt used to generate structured triplets from free-text reports.}
\vspace{-1em}
\label{prompt}
\end{table}

\begin{figure}[!t]
\centerline{\includegraphics[width=\textwidth]{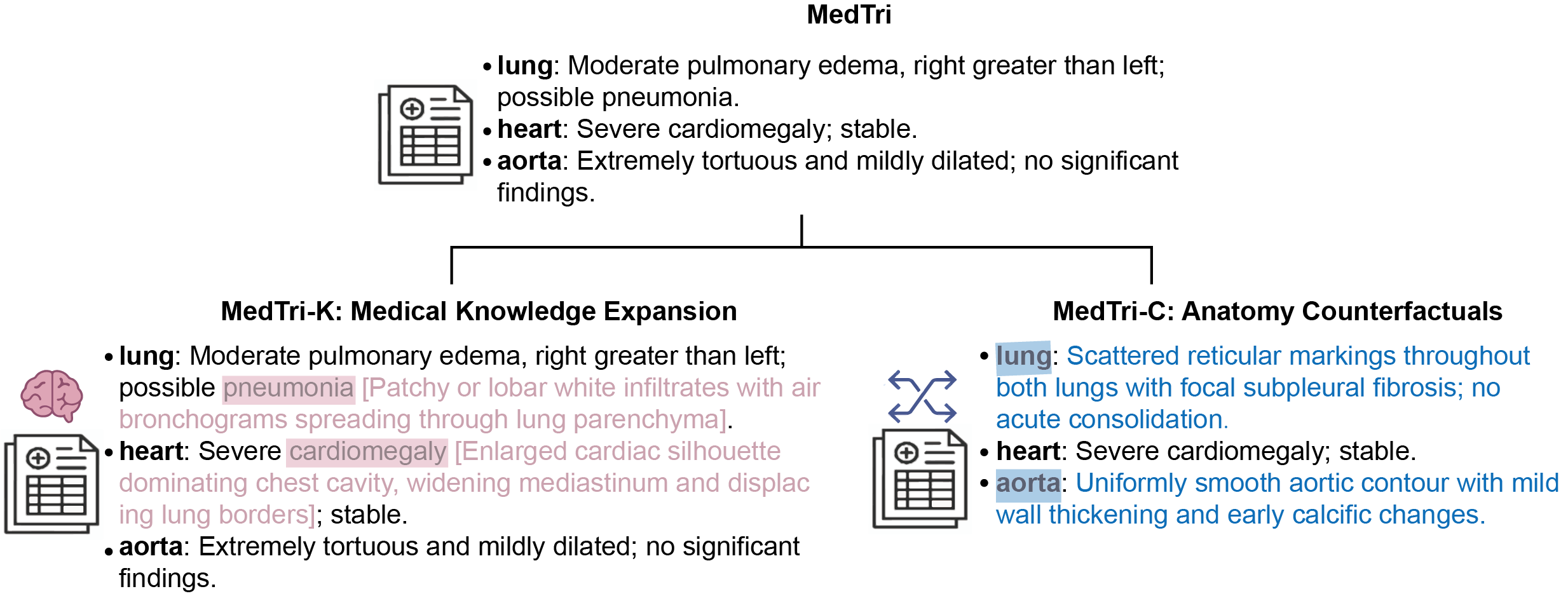}}
	\caption{Illustration of MedTri’s optional augmentation modules. MedTri-K enriches each triplet with clinically validated radiologic signatures to improve visual interpretability. MedTri-C generates anatomically inconsistent counterfactuals through controlled perturbations to strengthen fine-grained visual–semantic discrimination.}
 \vspace{-1.5em}
	\label{augment}
\end{figure}

\subsection{Optional Text-Level Augmentation on MedTri}
\subsubsection{Medical Knowledge Expansion (MedTri-K)}
Prior work in vision–language learning has shown that explicitly grounding diagnostic terms in visual attributes can improve semantic alignment~\cite{medfilip}. However, such strategies can be difficult to apply reliably to raw radiology reports due to stylistic variability and entangled narrative structure. Leveraging the structured triplet produced by MedTri, we can integrate a lightweight knowledge expansion mechanism (Fig. 2, left). For each normalized triplet, the diagnosis is augmented with a concise description of its characteristic radiological appearance, retrieved from our created dictionary of standard medical definitions covering over 100 common radiological findings and diagnoses. For example, pneumonia is associated with parenchymal consolidation or high-attenuation opacity within the affected lobe. These dictionary entries are created by board-certified radiologists and normalized for terminological consistency, allowing them to be seamlessly integrated into MedTri without altering its underlying structure. Importantly, this module operates entirely on normalized text and does not modify the core normalization process.

\subsubsection{Anatomy-Grounded Counterfactuals (MedTri-C)}
Counterfactual supervision and hard negative sampling have been widely explored to encourage fine-grained discrimination in vision–language models~\cite{counter_1}\cite{counter_2}\cite{Feng_2023_CVPR}. The anatomically grounded triplet structure produced by MedTri enables a controlled and fine-grained instantiation of counterfactual text augmentation through localized perturbations.Specifically, we generate counterfactual reports by modifying the descriptions of several anatomical entities (n=2 in our study) at a time, replacing them with semantically incompatible counterparts randomly sampled from other normalized triplets (Fig 2, right). Replacement is constrained within the same anatomical hierarchy level to ensure structural consistency while altering local semantic alignment. These substitutions preserve the overall syntactic format and global report semantics, while deliberately disrupting the local factual alignment between anatomy and imaging findings.

The resulting counterfactual texts are paired with the original images and treated as hard negatives during contrastive training. By introducing fine-grained, anatomy-level inconsistencies rather than global semantic shifts, this strategy forces the model to attend to localized visual evidence and fine-grained anatomical features, instead of relying on coarse diagnostic signals. As with MedTri-K, this module is optional and applied only during training, without affecting the normalization backbone.

\section{Experiments and Results}
\vspace{-0.75em}
\subsection{Evaluation of MedTri for Normalization}
To assess the quality and deployability of the proposed platform, we compare MedTri against two representative baselines: (1) ChatGPT-5.1-based structured rewriting, which serves as a high-quality but non-deployable reference for distilled supervision, and (2) Qwen2.5-14B~\cite{qwen}, a compact open-source model commonly used in local summarization workflows. Quantitative results are summarized in Table~\ref{eva}.

To evaluate clinical validity beyond surface-level textual similarity and to mitigate potential reference bias, we conduct a physician expert evaluation as the primary assessment. Twenty board-certified physicians independently assess the normalized reports in a double-blind manner using a five-point Likert scale, focusing on anatomical correctness and image groundedness. As shown in Table~\ref{eva}, MedTri achieves expert scores comparable to the cloud-based LLM reference and substantially higher than the open-source baseline, indicating that it preserves clinically meaningful and image-grounded information while remaining deployable in typical research and clinical settings.

We also report automatic text similarity metrics, including BERT score, BLEU, and ROUGE~\cite{feng2025noisy}\cite{text_pre_3}\cite{text_pre_4}, computed against ChatGPT-5.1-generated references. These metrics primarily measure the degree to which MedTri approximates the distilled reference normalization and are therefore used as proxy indicators of consistency rather than as direct measures of clinical correctness. Quantitative results for the open-source baseline are included for reference. 

\begin{table}
\vspace{-1em}
\centering
\label{tab:medtri_eval_inline}
\resizebox{0.7\textwidth}{!}{
\normalsize
\begin{tabular}{lccc}
\hline
\textbf{Metric} & \textbf{ChatGPT-5.1} & \textbf{Qwen2.5} & \textbf{MedTri} \\
\hline
\rowcolor[HTML]{DCDDDD}
Anatomical correctness (1-5) & 4.888 & 3.363     & 4.838 \\
\rowcolor[HTML]{DCDDDD}
Image groundedness (1-5)    & 4.775 & 2.938     & 4.763 \\
\rowcolor[HTML]{F7F8F8}
VRAM usage (GB)              & --          & 15.04       & 1.50 \\
\rowcolor[HTML]{F7F8F8}
Time/report (s)        & 3.25±2.14   & 1.21±0.67   & 0.45±0.18 \\
BERT score             & --          & 0.734±0.073  & 0.906±0.072 \\
BLEU                   & --          & 0.097       & 0.541 \\
ROUGE-1                & --          & 0.566±0.127  & 0.824±0.146 \\
ROUGE-2                & --          & 0.307±0.122 & 0.720±0.190 \\
\hline
\end{tabular}}
\caption{Comparison of computational efficiency, expert evaluation, and normalization accuracy across different systems.}
\vspace{-1em}
\label{eva}
\end{table}

\subsection{MedTri for Improving Downstream Tasks}
\subsubsection{Experiment Settings and Datasets}
We adopt Swin Transformer and Vision Transformer (ViT)~\cite{monai}\cite{frepa} as the image encoders, and BiomedVLP-CXR~\cite{BiomedVLP} as the text encoder. All reports were truncated to a fixed maximum length of 512 tokens, which sufficiently covers the majority of radiology reports~\cite{CTRATE} in our datasets and avoids disproportionately truncating raw reports, ensuring a fair comparison. Model training is conducted using the InfoNCE objective for contrastive learning. We apply data augmentation, including random horizontal flipping and Gaussian noise injection (sigma=0.05). All experiments are executed on an Ubuntu workstation equipped with a single NVIDIA A6000 GPU.

Pretraining is conducted separately for two imaging modalities. For 2D radiographs, we use the MIMIC-CXR dataset~\cite{mimic_xray} (n=370k), while for 3D volumetric imaging, we adopt the CT-RATE dataset~\cite{CTRATE} (n=42k) for CT pretraining.

For X-ray downstream evaluation, we conduct multi-label classification on three datasets: MIMIC-CXR, NIH ChestX-ray14~\cite{NIH} (n=112k), and RSNA-Pneumonia~\cite{Pneumonia} (n=30k). From each dataset, we randomly sample 1,024 studies for evaluation. 

CT downstream tasks are evaluated on the CT-RATE dataset, with the training and testing splits following the official protocol described in~\cite{CTRATE}. Due to the severe class imbalance commonly observed in medical imaging datasets, we determine the decision threshold that maximizes the F1 score and report the corresponding accuracy. Both F1 score and accuracy are computed on a per-label basis and then macro-averaged across all labels~\cite{medfilip}.

\subsubsection{Downstream Task Performances}
\begin{table}
\vspace{-2em}
\centering
\resizebox{\textwidth}{!}{
\begin{tabular}{lcccccccccccccccccc}
\hline
\multicolumn{1}{|l|}{\multirow{3}{*}{SwinT}} & \multicolumn{6}{c|}{MIMIC}                                                                                                                                                                                                        & \multicolumn{6}{c|}{NIH}                                                                                                                                                                                                         & \multicolumn{6}{c|}{Pneumonia}                                                                                                                                                                                                    \\ \cline{2-19} 
\multicolumn{1}{|l|}{}                       & \multicolumn{2}{c|}{1\%}                                                  & \multicolumn{2}{c|}{10\%}                                                 & \multicolumn{2}{c|}{100\%}                                                & \multicolumn{2}{c|}{1\%}                                                  & \multicolumn{2}{c|}{10\%}                                                 & \multicolumn{2}{c|}{100\%}                                                & \multicolumn{2}{c|}{1\%}                                                  & \multicolumn{2}{c|}{10\%}                                                 & \multicolumn{2}{c|}{100\%}                                                \\ \cline{2-19} 
\multicolumn{1}{|l|}{}                       & \multicolumn{1}{c|}{ACC}            & \multicolumn{1}{c|}{F1}             & \multicolumn{1}{c|}{ACC}            & \multicolumn{1}{c|}{F1}             & \multicolumn{1}{c|}{ACC}            & \multicolumn{1}{c|}{F1}             & \multicolumn{1}{c|}{ACC}            & \multicolumn{1}{c|}{F1}             & \multicolumn{1}{c|}{ACC}            & \multicolumn{1}{c|}{F1}             & \multicolumn{1}{c|}{ACC}            & \multicolumn{1}{c|}{F1}             & \multicolumn{1}{c|}{ACC}            & \multicolumn{1}{c|}{F1}             & \multicolumn{1}{c|}{ACC}            & \multicolumn{1}{c|}{F1}             & \multicolumn{1}{c|}{ACC}            & \multicolumn{1}{c|}{F1}             \\ \hline
\multicolumn{1}{|l|}{Raw report}             & \multicolumn{1}{c|}{0.612}          & \multicolumn{1}{c|}{0.218}          & \multicolumn{1}{c|}{0.701}          & \multicolumn{1}{c|}{0.263}          & \multicolumn{1}{c|}{0.781}          & \multicolumn{1}{c|}{0.302}          & \multicolumn{1}{c|}{0.655}          & \multicolumn{1}{c|}{0.221}          & \multicolumn{1}{c|}{0.722}          & \multicolumn{1}{c|}{0.266}          & \multicolumn{1}{c|}{0.762}          & \multicolumn{1}{c|}{0.293}          & \multicolumn{1}{c|}{0.812}          & \multicolumn{1}{c|}{0.673}          & \multicolumn{1}{c|}{0.833}          & \multicolumn{1}{c|}{0.699}          & \multicolumn{1}{c|}{0.848}          & \multicolumn{1}{c|}{0.722}          \\ \hline
\multicolumn{1}{|l|}{RadGraph}               & \multicolumn{1}{c|}{0.584}          & \multicolumn{1}{c|}{0.205}          & \multicolumn{1}{c|}{0.673}          & \multicolumn{1}{c|}{0.251}          & \multicolumn{1}{c|}{0.748}          & \multicolumn{1}{c|}{0.284}          & \multicolumn{1}{c|}{0.628}          & \multicolumn{1}{c|}{0.207}          & \multicolumn{1}{c|}{0.703}          & \multicolumn{1}{c|}{0.249}          & \multicolumn{1}{c|}{0.747}          & \multicolumn{1}{c|}{0.265}          & \multicolumn{1}{c|}{0.804}          & \multicolumn{1}{c|}{0.665}          & \multicolumn{1}{c|}{0.822}          & \multicolumn{1}{c|}{0.690}          & \multicolumn{1}{c|}{0.840}          & \multicolumn{1}{c|}{0.715}          \\ \hline
\multicolumn{1}{|l|}{MedTri}                 & \multicolumn{1}{c|}{0.628}          & \multicolumn{1}{c|}{0.229}          & \multicolumn{1}{c|}{0.716}          & \multicolumn{1}{c|}{0.279}          & \multicolumn{1}{c|}{0.792}          & \multicolumn{1}{c|}{0.314}          & \multicolumn{1}{c|}{0.672}          & \multicolumn{1}{c|}{0.233}          & \multicolumn{1}{c|}{0.739}          & \multicolumn{1}{c|}{0.281}          & \multicolumn{1}{c|}{0.779}          & \multicolumn{1}{c|}{0.304}          & \multicolumn{1}{c|}{0.817}          & \multicolumn{1}{c|}{0.676}          & \multicolumn{1}{c|}{0.836}          & \multicolumn{1}{c|}{0.701}          & \multicolumn{1}{c|}{0.850}          & \multicolumn{1}{c|}{0.723}          \\ \hline
\multicolumn{1}{|l|}{MedTri-K}               & \multicolumn{1}{c|}{\textbf{0.645}} & \multicolumn{1}{c|}{\textbf{0.241}} & \multicolumn{1}{c|}{\textbf{0.732}} & \multicolumn{1}{c|}{0.286}          & \multicolumn{1}{c|}{0.784}          & \multicolumn{1}{c|}{0.305}          & \multicolumn{1}{c|}{\textbf{0.689}} & \multicolumn{1}{c|}{\textbf{0.242}} & \multicolumn{1}{c|}{0.746}          & \multicolumn{1}{c|}{0.284}          & \multicolumn{1}{c|}{0.771}          & \multicolumn{1}{c|}{0.292}          & \multicolumn{1}{c|}{0.823}          & \multicolumn{1}{c|}{\textbf{0.681}} & \multicolumn{1}{c|}{\textbf{0.840}} & \multicolumn{1}{c|}{\textbf{0.703}} & \multicolumn{1}{c|}{0.851}          & \multicolumn{1}{c|}{0.722}          \\ \hline
\multicolumn{1}{|l|}{MedTri-C}               & \multicolumn{1}{c|}{0.629}          & \multicolumn{1}{c|}{0.226}          & \multicolumn{1}{c|}{0.731}          & \multicolumn{1}{c|}{\textbf{0.291}} & \multicolumn{1}{c|}{\textbf{0.803}} & \multicolumn{1}{c|}{\textbf{0.318}} & \multicolumn{1}{c|}{0.674}          & \multicolumn{1}{c|}{0.227}          & \multicolumn{1}{c|}{\textbf{0.751}} & \multicolumn{1}{c|}{\textbf{0.290}} & \multicolumn{1}{c|}{\textbf{0.801}} & \multicolumn{1}{c|}{\textbf{0.321}} & \multicolumn{1}{c|}{\textbf{0.824}} & \multicolumn{1}{c|}{0.679}          & \multicolumn{1}{c|}{0.838}          & \multicolumn{1}{c|}{0.701}          & \multicolumn{1}{c|}{\textbf{0.855}} & \multicolumn{1}{c|}{\textbf{0.728}} \\ \hline
                                             &                                     &                                     &                                     &                                     &                                     &                                     &                                     &                                     &                                     &                                     &                                     &                                     &                                     &                                     &                                     &                                     &                                     &                                     \\ \hline
\multicolumn{1}{|l|}{\multirow{3}{*}{ViT}}   & \multicolumn{6}{c|}{MIMIC}                                                                                                                                                                                                        & \multicolumn{6}{c|}{NIH}                                                                                                                                                                                                         & \multicolumn{6}{c|}{Pneumonia}                                                                                                                                                                                                    \\ \cline{2-19} 
\multicolumn{1}{|l|}{}                       & \multicolumn{2}{c|}{1\%}                                                  & \multicolumn{2}{c|}{10\%}                                                 & \multicolumn{2}{c|}{100\%}                                                & \multicolumn{2}{c|}{1\%}                                                  & \multicolumn{2}{c|}{10\%}                                                 & \multicolumn{2}{c|}{100\%}                                                & \multicolumn{2}{c|}{1\%}                                                  & \multicolumn{2}{c|}{10\%}                                                 & \multicolumn{2}{c|}{100\%}                                                \\ \cline{2-19} 
\multicolumn{1}{|l|}{}                       & \multicolumn{1}{c|}{ACC}            & \multicolumn{1}{c|}{F1}             & \multicolumn{1}{c|}{ACC}            & \multicolumn{1}{c|}{F1}             & \multicolumn{1}{c|}{ACC}            & \multicolumn{1}{c|}{F1}             & \multicolumn{1}{c|}{ACC}            & \multicolumn{1}{c|}{F1}             & \multicolumn{1}{c|}{ACC}            & \multicolumn{1}{c|}{F1}             & \multicolumn{1}{c|}{ACC}            & \multicolumn{1}{c|}{F1}             & \multicolumn{1}{c|}{ACC}            & \multicolumn{1}{c|}{F1}             & \multicolumn{1}{c|}{ACC}            & \multicolumn{1}{c|}{F1}             & \multicolumn{1}{c|}{ACC}            & \multicolumn{1}{c|}{F1}             \\ \hline
\multicolumn{1}{|l|}{Raw report}             & \multicolumn{1}{c|}{0.609}          & \multicolumn{1}{c|}{0.215}          & \multicolumn{1}{c|}{0.703}          & \multicolumn{1}{c|}{0.265}          & \multicolumn{1}{c|}{0.774}          & \multicolumn{1}{c|}{0.282}          & \multicolumn{1}{c|}{0.655}          & \multicolumn{1}{c|}{0.221}          & \multicolumn{1}{c|}{0.720}          & \multicolumn{1}{c|}{0.262}          & \multicolumn{1}{c|}{0.787}          & \multicolumn{1}{c|}{0.284}          & \multicolumn{1}{c|}{0.794}          & \multicolumn{1}{c|}{0.644}          & \multicolumn{1}{c|}{0.827}          & \multicolumn{1}{c|}{0.683}          & \multicolumn{1}{c|}{0.847}          & \multicolumn{1}{c|}{0.718}          \\ \hline
\multicolumn{1}{|l|}{RadGraph}               & \multicolumn{1}{c|}{0.587}          & \multicolumn{1}{c|}{0.207}          & \multicolumn{1}{c|}{0.674}          & \multicolumn{1}{c|}{0.250}          & \multicolumn{1}{c|}{0.741}          & \multicolumn{1}{c|}{0.262}          & \multicolumn{1}{c|}{0.631}          & \multicolumn{1}{c|}{0.209}          & \multicolumn{1}{c|}{0.702}          & \multicolumn{1}{c|}{0.248}          & \multicolumn{1}{c|}{0.738}          & \multicolumn{1}{c|}{0.266}          & \multicolumn{1}{c|}{0.785}          & \multicolumn{1}{c|}{0.649}          & \multicolumn{1}{c|}{0.813}          & \multicolumn{1}{c|}{0.678}          & \multicolumn{1}{c|}{0.834}          & \multicolumn{1}{c|}{0.707}          \\ \hline
\multicolumn{1}{|l|}{MedTri}                 & \multicolumn{1}{c|}{0.629}          & \multicolumn{1}{c|}{0.229}          & \multicolumn{1}{c|}{0.717}          & \multicolumn{1}{c|}{0.280}          & \multicolumn{1}{c|}{0.787}          & \multicolumn{1}{c|}{0.299}          & \multicolumn{1}{c|}{0.673}          & \multicolumn{1}{c|}{0.235}          & \multicolumn{1}{c|}{0.738}          & \multicolumn{1}{c|}{0.281}          & \multicolumn{1}{c|}{0.792}          & \multicolumn{1}{c|}{0.306}          & \multicolumn{1}{c|}{0.801}          & \multicolumn{1}{c|}{0.660}          & \multicolumn{1}{c|}{0.832}          & \multicolumn{1}{c|}{0.693}          & \multicolumn{1}{c|}{0.846}          & \multicolumn{1}{c|}{0.721}          \\ \hline
\multicolumn{1}{|l|}{MedTri-K}               & \multicolumn{1}{c|}{\textbf{0.647}} & \multicolumn{1}{c|}{\textbf{0.241}} & \multicolumn{1}{c|}{\textbf{0.734}} & \multicolumn{1}{c|}{0.288}          & \multicolumn{1}{c|}{0.782}          & \multicolumn{1}{c|}{0.289}          & \multicolumn{1}{c|}{\textbf{0.690}} & \multicolumn{1}{c|}{\textbf{0.244}} & \multicolumn{1}{c|}{\textbf{0.746}} & \multicolumn{1}{c|}{0.285}          & \multicolumn{1}{c|}{0.792}          & \multicolumn{1}{c|}{0.306}          & \multicolumn{1}{c|}{\textbf{0.806}} & \multicolumn{1}{c|}{\textbf{0.671}} & \multicolumn{1}{c|}{0.841}          & \multicolumn{1}{c|}{\textbf{0.706}} & \multicolumn{1}{c|}{0.842}          & \multicolumn{1}{c|}{0.723}          \\ \hline
\multicolumn{1}{|l|}{MedTri-C}               & \multicolumn{1}{c|}{0.625}          & \multicolumn{1}{c|}{0.226}          & \multicolumn{1}{c|}{\textbf{0.734}} & \multicolumn{1}{c|}{\textbf{0.291}} & \multicolumn{1}{c|}{\textbf{0.801}} & \multicolumn{1}{c|}{\textbf{0.307}} & \multicolumn{1}{c|}{0.677}          & \multicolumn{1}{c|}{0.238}          & \multicolumn{1}{c|}{0.744}          & \multicolumn{1}{c|}{\textbf{0.289}} & \multicolumn{1}{c|}{\textbf{0.808}} & \multicolumn{1}{c|}{\textbf{0.312}} & \multicolumn{1}{c|}{0.803}          & \multicolumn{1}{c|}{0.669}          & \multicolumn{1}{c|}{\textbf{0.846}} & \multicolumn{1}{c|}{0.704}          & \multicolumn{1}{c|}{\textbf{0.855}} & \multicolumn{1}{c|}{\textbf{0.728}} \\ \hline
\end{tabular}} 
\caption{Downstream classification performance of Swin Transformer (SwinT) and ViT on different X-ray datasets under different text preprocessing strategies. Best performance is marked in bold.}
\vspace{-1em}
\label{xray}
\end{table}
The quantitative results are presented in Table~\ref{xray} and \ref{ctrate}, respectively. Across all X-ray and CT benchmarks, MedTri and its variants consistently outperform raw reports and RadGraph across different data scales and visual backbones. The improvements are observed for both accuracy and F1 score, with particularly pronounced gains in low-data regimes (1\% and 10\%), indicating that structured normalization substantially improves sample efficiency for vision--language pretraining. The consistent performance gains across datasets and architectures demonstrate that the benefits of MedTri are robust and stem primarily from improved textual supervision.
\begin{table}
\vspace{-1em}
\centering
\resizebox{0.9\textwidth}{!}{
\normalsize
\begin{tabular}{|l|cccccc|cccccc|}
\hline
\multicolumn{1}{|c|}{\multirow{3}{*}{}} & \multicolumn{6}{c|}{SwinT}                                                                                                                                                                                   & \multicolumn{6}{c|}{ViT}                                                                                                                                                                                     \\ \cline{2-13} 
\multicolumn{1}{|c|}{}                  & \multicolumn{2}{c|}{1\%}                                                  & \multicolumn{2}{c|}{10\%}                                                 & \multicolumn{2}{c|}{100\%}                           & \multicolumn{2}{c|}{1\%}                                                  & \multicolumn{2}{c|}{10\%}                                                 & \multicolumn{2}{c|}{100\%}                           \\ \cline{2-13} 
\multicolumn{1}{|c|}{}                  & \multicolumn{1}{c|}{ACC}            & \multicolumn{1}{c|}{F1}             & \multicolumn{1}{c|}{ACC}            & \multicolumn{1}{c|}{F1}             & \multicolumn{1}{c|}{ACC}            & F1             & \multicolumn{1}{c|}{ACC}            & \multicolumn{1}{c|}{F1}             & \multicolumn{1}{c|}{ACC}            & \multicolumn{1}{c|}{F1}             & \multicolumn{1}{c|}{ACC}            & F1             \\ \hline
Raw report                              & \multicolumn{1}{c|}{0.719}          & \multicolumn{1}{c|}{0.493}          & \multicolumn{1}{c|}{0.741}          & \multicolumn{1}{c|}{0.518}          & \multicolumn{1}{c|}{0.775}          & 0.525          & \multicolumn{1}{c|}{0.696}          & \multicolumn{1}{c|}{0.473}          & \multicolumn{1}{c|}{0.733}          & \multicolumn{1}{c|}{0.511}          & \multicolumn{1}{c|}{0.745}          & 0.518          \\ \hline
RadGraph                                & \multicolumn{1}{c|}{0.714}          & \multicolumn{1}{c|}{0.480}          & \multicolumn{1}{c|}{0.735}          & \multicolumn{1}{c|}{0.513}          & \multicolumn{1}{c|}{0.744}          & 0.523          & \multicolumn{1}{c|}{0.694}          & \multicolumn{1}{c|}{0.471}          & \multicolumn{1}{c|}{0.722}          & \multicolumn{1}{c|}{0.508}          & \multicolumn{1}{c|}{0.716}          & 0.509          \\ \hline
MedTri                                  & \multicolumn{1}{c|}{0.723}          & \multicolumn{1}{c|}{0.499}          & \multicolumn{1}{c|}{0.753}          & \multicolumn{1}{c|}{0.520}          & \multicolumn{1}{c|}{0.785}          & 0.538          & \multicolumn{1}{c|}{0.704}          & \multicolumn{1}{c|}{0.488}          & \multicolumn{1}{c|}{0.745}          & \multicolumn{1}{c|}{0.518}          & \multicolumn{1}{c|}{0.775}          & 0.537          \\ \hline
MedTri-K                                & \multicolumn{1}{c|}{\textbf{0.732}} & \multicolumn{1}{c|}{\textbf{0.510}} & \multicolumn{1}{c|}{0.757}          & \multicolumn{1}{c|}{0.519}          & \multicolumn{1}{c|}{0.789}          & 0.536          & \multicolumn{1}{c|}{\textbf{0.718}} & \multicolumn{1}{c|}{\textbf{0.498}} & \multicolumn{1}{c|}{0.743}          & \multicolumn{1}{c|}{\textbf{0.519}} & \multicolumn{1}{c|}{0.775}          & 0.530          \\ \hline
MedTri-C                                & \multicolumn{1}{c|}{0.726}          & \multicolumn{1}{c|}{0.501}          & \multicolumn{1}{c|}{\textbf{0.767}} & \multicolumn{1}{c|}{\textbf{0.523}} & \multicolumn{1}{c|}{\textbf{0.794}} & \textbf{0.557} & \multicolumn{1}{c|}{0.704}          & \multicolumn{1}{c|}{0.485}          & \multicolumn{1}{c|}{\textbf{0.751}} & \multicolumn{1}{c|}{\textbf{0.519}} & \multicolumn{1}{c|}{\textbf{0.782}} & \textbf{0.543} \\ \hline
\end{tabular}} 
\caption{Downstream classification performance of Swin Transformer (SwinT) and ViT on CT-RATE datasets under different text preprocessing strategies. Best performance is marked in bold.}
\vspace{-1em}
\label{ctrate}
\end{table}
The two optional augmentation modules exhibit complementary behaviors. MedTri-K (knowledge expansion) tends to achieve the best or near-best performance under limited data settings (1\% and 10\%), suggesting that explicitly linking diagnostic terms to characteristic imaging appearances provides additional semantic grounding when training data is scarce. However, its gains diminish at full-data scale (100\%), where the model can already learn such associations implicitly from abundant image--text pairs. In contrast, MedTri-C (counterfactual construction) shows limited improvement in the 1\% setting, likely because extremely limited data prevents the model from effectively exploiting fine-grained counterfactual distinctions. As data availability increases (10\% and 100\%), counterfactual supervision becomes more effective, leading to stronger gains in medium- and full-data regimes by encouraging finer-grained visual-semantic discrimination.

\section{Discussion and Conclusion}
\vspace{-0.75em}
In this work, we present MedTri, a lightweight and deployable medical report normalization platform that converts free-text medical reports into structured, anatomically grounded triplets for vision-language pretraining. Through extensive experiments on both X-ray and CT datasets, we demonstrate that structured normalization alone is an important factor in improving downstream performance, yielding consistent gains across different data scales, datasets, and visual backbones. The proposed platform provides an effective and practical alternative to cloud LLM-dependent pipelines, enabling scalable and privacy-preserving deployment in clinical and research settings.

Despite its effectiveness, this study has several limitations. First, our evaluation focuses on CLIP-style contrastive vision--language pretraining, which is widely adopted and representative, but does not cover other training paradigms such as generative, instruction-tuned, or task-specific multimodal learning frameworks. Second, MedTri is currently evaluated only on radiology reports and imaging modalities, and its applicability to other medical domains or non-radiological clinical narratives remains unexplored. In addition, the proposed triplet schema represents one principled design choice for structured normalization, and alternative schema formulations or decomposition strategies may need further comparative investigation. Addressing these limitations constitutes an important direction for future work.

\bibliographystyle{splncs04}
\bibliography{ref}
\end{document}